\documentclass[conference]{IEEEtran}
\IEEEoverridecommandlockouts
\usepackage{cite}
\usepackage{amsmath,amssymb,amsfonts}
\usepackage{algorithmic}
\usepackage{graphicx}
\usepackage{textcomp}
\usepackage{xcolor}
\usepackage{multirow}
\usepackage{booktabs}
\usepackage{threeparttable}
\usepackage[ruled,norelsize,vlined,linesnumbered]{algorithm2e}

\makeatletter
\newcommand{\removelatexerror}{\let\@latex@error\@gobble}
\makeatother
\begin{document}

\title{End-to-End Graph Flattening Method for Large Language Models}
\author{\IEEEauthorblockN{Bin Hong}
\IEEEauthorblockA{\textit{State Key Laboratory of Cognitive Intelligence}\\
\textit{University of Science and Technology of China}\\
Hefei, China \\
hb2002@mail.ustc.edu.cn}
\and
\IEEEauthorblockN{Jinze Wu}
\IEEEauthorblockA{\textit{iFLYTEK Research,} \\
\textit{iFLYTEK Co., Ltd}\\
Hefei, China \\
jzwu4@iflytek.com}
\and
\IEEEauthorblockN{Jiayu Liu}
\IEEEauthorblockA{\textit{State Key Laboratory of Cognitive Intelligence}\\
\textit{University of Science and Technology of China}\\
Hefei, China \\
jy251198@mail.ustc.edu.cn}
\and
\IEEEauthorblockN{Liang Ding}
\IEEEauthorblockA{\textit{iFLYTEK Research,} \\
\textit{iFLYTEK Co., Ltd}\\
Hefei, China \\
liangding3@iflytek.com}
\and
\IEEEauthorblockN{Jing Sha}
\IEEEauthorblockA{\textit{iFLYTEK Research,} \\
\textit{iFLYTEK Co., Ltd}\\
Hefei, China \\
jingsha@iflytek.com}
\and
\IEEEauthorblockN{Kai Zhang*}\thanks{*Kai Zhang is the corresponding author.}
\IEEEauthorblockA{\textit{State Key Laboratory of Cognitive Intelligence}\\
\textit{University of Science and Technology of China}\\
Hefei, China \\
kkzhang08@ustc.edu.cn}
\and
\IEEEauthorblockN{Shijin Wang}
\IEEEauthorblockA{\textit{iFLYTEK AI Research (Central China),}\\
\textit{iFLYTEK Co., Ltd}\\
Hefei, China \\
sjwang3@iflytek.com}
\and
\IEEEauthorblockN{Zhenya Huang}
\IEEEauthorblockA{\textit{State Key Laboratory of Cognitive Intelligence}\\
\textit{University of Science and Technology of China}\\
Hefei, China \\
huangzhy@ustc.edu.cn}
}

\maketitle

\begin{abstract}
In recent years, the breakthrough of Large Language Models (LLMs) offers new ideas for achieving universal methods on graph data. The common practice of converting graphs into natural language for LLMs, which refers to graph flattening, exhibits good generalizability and interpretability. However, the poor organization of the textual format results in poor performance in long-distance scenario understanding. Inspired by human cognitive reasoning habits, we propose a novel method for graph flattening to fit LLMs, termed as End-to-End DAG-Path prompting (EEDP). Experiments on real-world datasets show that EEDP enhances the reasoning performance of LLMs in long-distance scenarios while maintaining excellent performance in short-distance scenarios, demonstrating good robustness in the face of distance variations.
\end{abstract}


\section{Introduction}

Recent progress in large language models (LLMs) has been remarkable, showcasing strong text processing capabilities across various tasks, including those involving multi-modal data such as graphs \cite{zhao2023survey}. Graphs are crucial in applications like social networks \cite{yang2021consisrec}, recommender systems \cite{wang2021privileged}, knowledge graphs \cite{liu2023enhancing,liu2023rhgn}, molecular graphs \cite{guo2021few}, and natural language processing \cite{zhang2024label,zhang2022graph}. Leveraging LLMs for graph data aims to achieve a universal method for graph processing \cite{chen2024exploring}.

LLMs have been applied to tasks with implicit graph structures, such as path planning \cite{tandon2019wiqa} and multi-hop question answering \cite{madaan2022language}. Recent research has explored LLMs' ability to understand explicit graph structures \cite{wang2024can,fatemi2023talk}. Since LLMs cannot directly handle explicit graphs, they use textual descriptions, known as graph-flattening \cite{li2023survey}, like adjacency lists. Flatten-based methods fit various downstream tasks, demonstrating generalizability and interpretability.

Many real-world graph-related tasks require handling long-range dependencies, such as textual coherence analysis, inference chains, and time series forecasting. Existing flattened methods are effective in short-distance scenarios but perform poorly in long-distance ones. The textual format of a flattened graph significantly influences LLM performance \cite{guo2023gpt4graph}. Thus, designing an efficient flattened format for graphs is necessary.

We explores optimizing graph-flattening by combining human cognitive habits and real-world graph structures. We propose a novel graph representation method, \textit{\textbf{E}nd-to-\textbf{E}nd \textbf{DAG-\textbf{P}ath (EEDP) prompting}}, which leverages main backbone paths within a graph to generate textual descriptions. Wang et al. \cite{wang2024can} found learning from examples did not occur in complex graph reasoning problems, so we focus on zero-shot performance. We establish two benchmarks based on real-world data: Merged\_1000 and ZINC\_test\_2500. Experiments demonstrate our method's effectiveness in both long-distance and short-distance scenarios.

\section{Related Work}

\begin{figure*}[htbp]
\centerline{\includegraphics[width=0.95\textwidth]{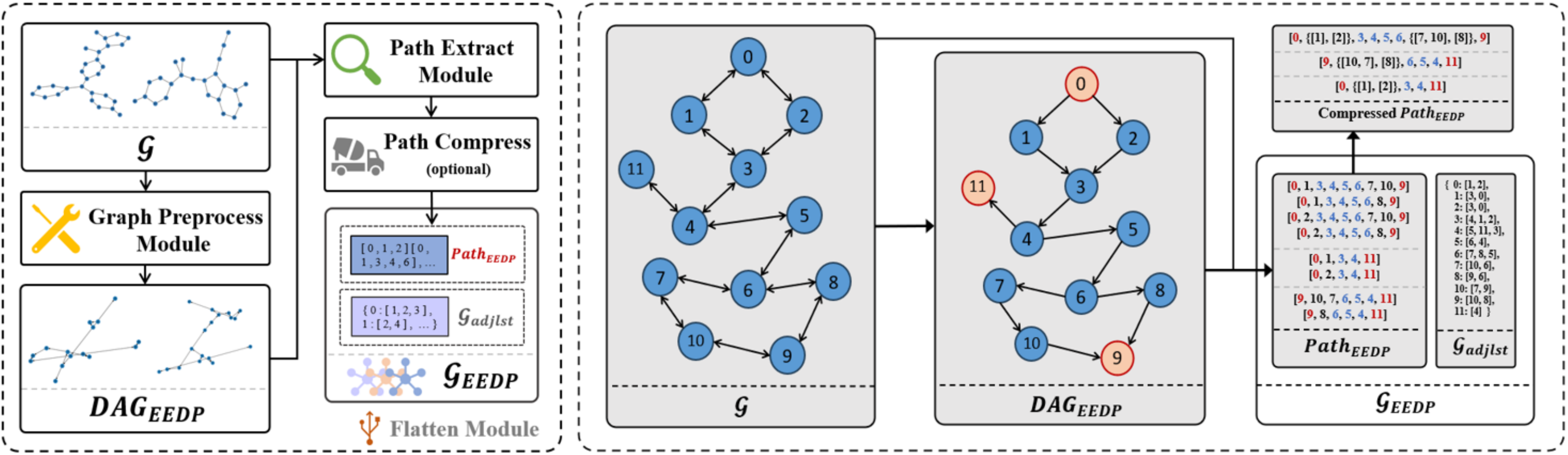}}
\caption{The framework of EEDP and an example of how EEDP handles a input graph $\mathcal{G}$. It is first used to generate $DAG_{\text{EEDP}}$. $Path_{\text{EEDP}}$ are extracted from $\mathcal{G}$ based on the endpoints $\mathcal{V}_{\text{end}}$ found by using $DAG_{\text{EEDP}}$. $Path_{\text{EEDP}}$ can be compressed if needed. Finally, $Path_{\text{EEDP}}$ and $\mathcal{G}_{\text{adjlst}}$ are concatenated to obtain the EEDP-flattened graph $\mathcal{G}_{\text{EEDP}}$.}
\label{Fig. 2.}
\end{figure*}

\subsection{Graph Description Language}
Graph description languages are standardized languages used to define or represent graph-structured data \cite{guo2023gpt4graph}. Himsolt et al. proposed a Python-like language called Graph Modelling Language (GML) \cite{himsolt1997gml}, while Brandes et al. introduced an XML-based language called Graph Markup Language (GraphML) \cite{brandes2013graph}. In a broader sense, graph description languages include various traditional representations based on 1-hop connections, such as adjacency lists, adjacency matrices, and edge lists.

In graph deep learning, processing large-scale graph structures often requires specialized methods. Some approaches employ walk sequences. RandomWalk \cite{lawler2010random} starts from a node and randomly jumps to a neighbor, generating a sequence after several hops, which is then used to create new node representations. Node2VecWalk \cite{grover2016node2vec} optimizes RandomWalk by weighting transition probabilities. GraphSAGE \cite{hamilton2017inductive} constructs an ego-graph for each node, starting from a central node and sampling neighbors within a k-hop radius.

\section{Methodology}
\subsection{Problem Definition}

Graph-flattening method generates a textual representation of the input graph $\mathcal{G} = \{\mathcal{V}, \mathcal{E}\}$: $\mathcal{G}_{\text{flat}} = f_{\text{flat}}(\mathcal{G})$. 

EEDP optimizes $f_{\text{flat}}$ similarly to human cognition by using the main backbone paths $Path_{\text{EEDP}}$ among endpoints $\mathcal{V}_{\text{end}}$ within $\mathcal{G}$ as the main components of the textual description. 

We define endpoints as nodes with either zero in-degree or zero out-degree and main backbone paths as the paths connecting each pair of these endpoints.

In our method, a special directed acyclic graph $\mathcal{DAG}_{\text{EEDP}}$ is generated based on $\mathcal{G}$. Then The definition of $\mathcal{V}_{\text{end}}$ can be formalized as below:

\begin{equation}
\begin{aligned}
\mathcal{V}_{\text{end}} = \{v \mid & \, v \in DAG_{\text{EEDP}}.\mathcal{V}, \, v.\text{in\_degree} = 0 \\
                        & \, \lor \, v.\text{out\_degree} = 0\}.
\end{aligned}
\end{equation}

For every simple path $p$ in $\mathcal{G}$, we use $p.\text{start}$ and $p.\text{end}$ to denote the start node and end node separately. 
Then the formal definition of $Path_{\text{EEDP}}$ is:

\begin{equation}
\begin{aligned}
    Path_{\text{EEDP}} = \{p \mid & \, p \in \mathcal{G}.\text{all\_simple\_paths()}, \\
                                  & \, p.\text{start} \in \mathcal{V}_{\text{end}} \, \land \, p.\text{end} \in \mathcal{V}_{\text{end}}\}
\end{aligned}
\end{equation}

For those paths in $Path_{\text{EEDP}}$ which exists in both $\mathcal{G}$ and $DAG_{\text{EEDP}}$, we define them as $Path_{\text{DAG}}$.

\subsection{End-to-End DAG-Path Prompting}

To build a flattened-graph $\mathcal{G}_{\text{EEDP}}$ upon $\mathcal{V}_{\text{end}}$ and $Path_{\text{EEDP}}$, we introduce our EEDP framework, as shown in Fig. \ref{Fig. 2.}. 
To further enhance the capability of EEDP in short-distance scenarios, we concatenate $Path_{\text{EEDP}}$ together with $\mathcal{G}_{\text{adjlst}}$ to get the final textual representation $\mathcal{G}_{\text{EEDP}}$. An example of how our framework processes a graph is shown in Fig. \ref{Fig. 2.}.

\subsubsection{Graph Preprocessing Module}
To find the endpoints to generate the $Path_{\text{EEDP}}$, We firstly transfer the graph to a special directed acyclic graph $DAG_{\text{EEDP}}$. $DAG_{\text{EEDP}}$ has the properties as sharing the same set of nodes $\mathcal{V}$ as the input graph $\mathcal{G}$ and containing both nodes with zero in-degree and nodes with zero out-degree. 

These favorable properties of $DAG_{\text{EEDP}}$ ensure the existence of $Path_{\text{EEDP}}$. Therefore, generating $DAG_{\text{EEDP}}$ from the input $\mathcal{G}$ is the most crucial step in the EEDP method.

To obtain the desired $DAG_{\text{EEDP}}$ while minimizing information loss, we propose EEDP-DAG Algorithm based on BFS, as shown in Algorithm \ref{algo1}. In EEDP-DAG Algorithm, $DAG_{\text{EEDP}}$ is initialized as an empty directed graph. The algorithm traverses the edges of the input graph $\mathcal{G}$ in breadth-first order and selectively adds edges to $DAG_{\text{EEDP}}$ to avoid cycles. 

\begin{algorithm}[h]
  \SetAlgoLined
  \KwData{$\mathcal{G}$}
  \KwResult{$DAG_{\text{EEDP}}$}

  $search\_list \leftarrow \text{empty\_queue()}$\;
  $node \leftarrow \mathcal{G}.\mathcal{V}[0]$\;
  $visited\_heads \leftarrow \text{empty\_set()}$\;
  $future\_heads \leftarrow \text{empty\_set()}$\;
  $DAG_{\text{EEDP}} \leftarrow \text{empty\_DirectedGraph()}$\;
  \ForEach{$edge \in \mathcal{G}.\mathcal{E}$}{
    \If{$edge.\text{head} == node$}{
      $search\_list.\text{enqueue}(edge)$\;
    }
  }
  $future\_heads.\text{add}(node)$\;
  \While{$search\_list.\text{length} > 0$}{
    $current\_edge \leftarrow search\_list.\text{dequeue}()$\;
    $head \leftarrow current\_edge.\text{head}$\;
    $tail \leftarrow current\_edge.\text{tail}$\;
    \If{$tail \in visited\_heads$ \textbf{and} $tail \in future\_heads$}{
      \textbf{continue}\;
    }
    \If{$current\_edge.\text{reverse} \notin DAG_{EEDP}.\mathcal{E}$}{
      $DAG_{EEDP}.\text{add\_edge}(current\_edge)$\;
      $visited\_heads.\text{add}(head)$\;
    }
    $flag \leftarrow \text{False}$\;
    \ForEach{$edge \in \mathcal{G}.\mathcal{E}$}{
      \If{$edge.\text{head} == tail$ \textbf{and} $edge \notin DAG_{EEDP}.\mathcal{E}$ \textbf{and} $edge.\text{reverse} \notin DAG_{EEDP}.\mathcal{E}$}{
        $search\_list.\text{enqueue}(edge)$\;
        $flag \leftarrow \text{True}$\;
      }
    }
    \If{$flag$}{
      $future\_heads.\text{add}(tail)$\;
    }
  }
  
  \caption{EEDP-DAG Algorithm.}
  \label{algo1}
\end{algorithm}

\subsubsection{Path Extract Module}
In this part, we build the main part of EEDP upon endpoints.
As shown by Theorem 1, all the nodes in $\mathcal{V}_{\text{end}}$ of the generated $DAG_{\text{EEDP}}$ also exist in $\mathcal{G}.\mathcal{V}$. Therefore, Path Extract Module first uses the generated $DAG_{\text{EEDP}}$ to find the endpoints $\mathcal{V}_{\text{end}}$ in the input graph $\mathcal{G}$. After that, we can extract $Path_{\text{EEDP}}$ connecting each pair of the endpoints using depth-fist search (DFS).

To help LLMs better reasoning in short-distance scenarios, the adjacency list $\mathcal{G}_{\text{adjlst}}$ of the input $\mathcal{G}$ is introduced. For each head node in $\mathcal{G}$, we set it as a dictionary key and the adjacent tail nodes are grouped into a list and set as the dictionary value. Equation (2) outlines a formalized description of the construction process of $\mathcal{G}_{\text{adjlst}}$. Here is an example of the adjacency list:
$\{0: [1, 2, 3], 1: [2 ,3], 3: [2]\}$. 

The two representations of $\mathcal{G}$ are concatenated to obtain the final EEDP-flattened graph $\mathcal{G}_{\text{EEDP}}$, as shown in Fig. \ref{Fig. 2.}.

\subsubsection{Path Compress Module}
In the original $Path_{\text{EEDP}}$ between a endpoint pair, there are shared parts besides the endpoints. We utilize generalized lists to represent the different parts while merging the shared parts. There is an example of the compressed $\mathcal{G}_{\text{EEDP}}$ and the original $\mathcal{G}_{\text{EEDP}}$ on the right of Fig. \ref{Fig. 2.}. If needed, $\mathcal{G}_{\text{EEDP}}$ can be compressed in this manner to reduce the number of tokens in prompts. 

In particular, we design a differential pointer-based algorithm for path merging. We maintain a differential pointer for each path. The pointer advances only when pointing to a shared node; otherwise, only the pointer pointing to a non-shared node advances. While traversing the paths using differential pointers, we simultaneously track the shared segments and the non-shared segments of each path, and finally perform the merging. 

The Path Compressing Module reduces the average prompt length from 520.775 to 487.981 tokens on the Merged\_1000 dataset and from 2673.7248 to 1148.1972 tokens on the ZINC\_test\_2500 dataset\footnote{The numbers of tokens are calculated using the official tokenizer for ChatGPT: https://github.com/openai/tiktoken.}.

\section{Experiments}
\subsection{Task Definition}
To assess the model's understanding of graph structures, we designed two edge prediction tasks:

1. Edge Prediction - Connectivity Prediction (EP-CP): Given a pure graph and a pair of nodes, determine whether there is a directed path from the first node to the second node. The LLM must output a binary classification result, either "yes" or "no." The answer is considered correct if the model's response matches the ground truth label.

2. Edge Prediction - Distance Prediction (EP-DP): Given a pure graph and a pair of nodes, determine if there is a directed path from the source node to the target node. If such a path exists, the model must also output the length of the path. If no path exists, the path length is defined as -1. The answer is considered correct if it accurately reflects the length of any one of the simple paths between the given node pair in the graph.

\subsection{Benchmarks}
We construct two benchmarks from real-world graph data: Merged\_1000 and ZINC\_test\_2500. The statistics of them are listed in Table \uppercase\expandafter{\romannumeral1}. We used accuracy as the evaluation metric.

\begin{table*}[!t]
  \centering
  \caption{Statistics of benchmarks}
  \begin{tabular}{c c c c c c c c}
  \toprule
    \textbf{dataset} & \textbf{\# graphs}  & \textbf{\# avg. nodes}  & \textbf{\# avg. edges}  & \textbf{\# 1-hop} & \textbf{\# 2-hop}  & \textbf{\# 3-hop}  & \textbf{\# $\geq$5-hop} \\
    \midrule
    Merged\_1000 & 1000 & 13.16 & 12.11 & 2000 & 1822 & 1434 & 497  \\
    ZINC\_test\_2500 & 2500 & 23.07 & 49.60 & 10000 & 10000 & 10000 & 9995\\
    \bottomrule
  \end{tabular}
  \label{tab1}
\end{table*}

\begin{table*}[!t]
\belowrulesep=0pt
\aboverulesep=0pt
  \begin{center}
    \caption{Results on Connectivity Prediction task}
    \begin{tabular*}{\textwidth}{@{\extracolsep{\fill}} c |c c c c c | c c c c c}
    \toprule
    \multirow{2}{*}{\textbf{Methods}} & \multicolumn{5}{c|}{\textbf{Merged\_1000}} & \multicolumn{5}{c}{\textbf{ZINC\_test\_2500}}\\
    ~ & \textbf{1-hop} & \textbf{2-hop} & \textbf{3-hop} & \textbf{$\geq$5-hop} & \textbf{Total} & \textbf{1-hop} & \textbf{2-hop} & \textbf{3-hop} & \textbf{$\geq$5-hop} & \textbf{Total}\\
      \midrule
      Random & 50.00 & 50.00 & 50.00 & 50.00 & 50.00 & 50.00 & 50.00 & 50.00 & 50.00 & 50.00\\
      \midrule
      Adjacency Matrix & 96.20 & 63.34 & 42.33 & 30.38 & 66.68 & 96.53 & 82.41 & 73.25 & 56.53 & 77.18\\
      Adjacency List & 98.40 & 95.12 & 82.22 & 55.53 & 89.62 & 98.61 & 90.09 & 69.41 & 47.52 & 76.41\\
      Edge List & \textbf{99.45} & 93.30 & 84.45 & 56.34 & 90.04 & 99.30 & 85.37 & 64.05 & 44.36 & 73.27\\
      \midrule
      Ego-graph & 96.20 & 66.36 & 44.70 & 23.74 & 67.65 & 96.57 & 76.96 & 49.62 & 29.81 & 63.25\\
      Walk Sequence & 61.75 & 52.85 & 43.31 & 44.87 & 52.88 & 78.41 & 55.45 & 40.39 & 20.10 & 48.59\\
      \midrule
      GML & 99.40 & 91.22 & 75.03 & 39.64 & 85.57 & 99.01 & 79.83 & 53.41 & 29.94 & 65.55\\
      GraphML & 99.35 & 92.92 & 79.71 & 50.50 & 88.20 & 98.99 & 81.00 & 52.85 & 28.53 & 65.35\\
      \midrule
      Natural Language & 99.40 & 92.86 & 81.45 & 45.88 & 88.23 & 99.12 & 81.35 & 58.54 & 38.73 & 77.18\\
      \midrule
      EEDP & 99.05 & \textbf{97.53} & \textbf{96.09} & \textbf{92.15} & \textbf{97.24} & \textbf{99.40} & \textbf{91.44} & \textbf{87.55} & \textbf{82.07} & \textbf{90.12}\\
      \bottomrule
    \end{tabular*}
    \end{center}
      \label{tab2}
\end{table*}

\begin{table*}[!t]
\belowrulesep=0pt
\aboverulesep=0pt
  \begin{center}
    \caption{Results on Distance Prediction task}
    \begin{tabular*}{\textwidth}{@{\extracolsep{\fill}} c |c c c c c | c c c c c}
    \toprule
    \multirow{2}{*}{\textbf{Methods}} & \multicolumn{5}{c|}{\textbf{Merged\_1000}} & \multicolumn{5}{c}{\textbf{ZINC\_test\_2500}}\\
    ~ & \textbf{1-hop} & \textbf{2-hop} & \textbf{3-hop} & \textbf{$\geq$5-hop} & \textbf{Total} & \textbf{1-hop} & \textbf{2-hop} & \textbf{3-hop} & \textbf{$\geq$5-hop} & \textbf{Total}\\
      \midrule
      Random & 9.50 & 9.24 & 9.22 & 11.36 & 9.49 & 9.00 & 9.50 & 9.30 & 14.20 & 10.50\\
      \midrule
      Adjacency Matrix & 72.10 & 41.05 & 28.31 & 8.85 & 45.89 & 64.97 & 38.56 & 32.68 & 8.65 & 36.22\\
      Adjacency List & 86.40 & 68.50 & 57.53 & 33.20 & 68.94 & 90.55 & 60.05 & 49.98 & 17.58 & 54.54\\
      Edge List& 97.70 & 72.72 & 61.92 & 31.59 & 75.16 & 91.21 & 53.32 & 38.33 & 10.98 & 48.46\\
      \midrule
      Ego-graph & 90.65 & 62.51 & 47.07 & 17.30 & 64.54 & 82.99 & 58.51 & 41.34 & 9.43 & 48.07\\
      Walk Sequence & 58.60 & 33.10 & 23.43 & 13.08 & 37.82 & 52.98 & 24.27 & 17.27 & 2.60 & 24.28\\
      \midrule
      GML & \textbf{98.75} & 76.84 & 61.09 & 25.96 & 76.13 & 92.42 & \textbf{63.77} & 46.80 & 13.98 & 54.25\\
      GraphML & 98.10 & 78.49 & 65.97 & 24.55 & 77.52 & 86.91 & 60.06 & 45.96 & 13.22 & 51.54\\
      \midrule
      Natural Language & 98.35 & 78.27 & 68.27 & 34.81 & 79.00 & 93.47 & 57.13 & 43.33 & 13.52 & 51.87\\
      \midrule
      EEDP & 92.30 & \textbf{78.54} & \textbf{79.50} & \textbf{69.01} & \textbf{82.74} & \textbf{95.42} & 61.72 & \textbf{65.12} & \textbf{47.72} & \textbf{67.50}\\
      \bottomrule
    \end{tabular*}
    \end{center}
      \label{tab3}
\end{table*}

\subsubsection{Merged\_1000}
Merged\_1000 is a dataset built upon real-world educational data. We harvest pure graph structures from various educational knowledge graphs. Each graph represents a correlation between a range of knowledge concepts. The full Merged\_1000 dataset contains 1,000 graphs. 

\subsubsection{ZINC\_test\_2500}
We selected the publicly available real-world dataset ZINC \cite{bresson2019two}.
ZINC\footnote{https://www.chrsmrrs.com/graphkerneldatasets/ZINC\_test.zip} is a large molecular graph dataset widely used in computational chemistry. The official standard split of ZINC yielding a test set of 5,000 graphs. We randomly sampled 2,500 graphs from the ZINC\_test set.

For each graph, we sampled four groups of node pairs based on their shortest path distances: 1, 2, 3, and over 5 hops. We sampled four pairs of nodes for each distance category, resulting in a total of 16 node pairs per graph. If fewer than four pairs were available for a particular distance category, all available pairs were used. These graphs and test cases constitute the test set used in this study.

\subsection{Experimental Setup}
\subsubsection{LLM Backbone}
In our experiments, GPT-4-turbo\cite{achiam2023gpt} is selected as the backbone LLM. GPT-4-turbo is one of the most powerful LLMs, recognized for its high performance and widespread usage in research.

\subsubsection{Baseline}
We selected the following methods.

Traditional graph description languages: Adjacency Matrix, Adjacency List and Edge List.

Sampling methods for graph deep learning: Ego-graph, with neighbor range set to 1 and no limit for the number of neighbors. Walk Sequence, with the probability of transitioning set to the same and the upper limit of sequence length set to 5.

Structured graph description languages: Graph Modelling Language (GML)\cite{himsolt1997gml} and Graph Markup Language (GraphML)\cite{brandes2013graph}.

Others: Natural Language, which use natural language to describe edge lists. Random: For EP-CP task, answers are generated with equal probability from 'yes' and 'no'; for EP-DP task, a random integer is generated from -1 to the total number of nodes in the graph as the answer. This baseline is set to distinguish whether the LLM is performing non-random responsible reasoning in graph inference tasks.

\subsection{Results on Edge Prediction}

We conduct the prediction tasks as mentioned before. Table \uppercase\expandafter{\romannumeral2} and Table \uppercase\expandafter{\romannumeral3} report the overall results on both datasets.

From the tables, we can draw the following conclusions: 

\begin{table*}[!t]
\belowrulesep=0pt
\aboverulesep=0pt
  \centering
    \begin{threeparttable}
    \caption{Results of ablation study on EEDP}
    \begin{tabular*}{\textwidth}{@{\extracolsep{\fill}} c | c | c c c c c | c c c c c}
      \toprule
      \multirow{2}{*}{\textbf{Task}} & \multirow{2}{*}{\textbf{Method}} & \multicolumn{5}{c|}{\textbf{Merged\_1000}} & \multicolumn{5}{c}{\textbf{ZINC\_test\_2500}}\\
      ~ & ~ & \textbf{1-hop}  & \textbf{2-hop} & \textbf{3-hop} & \textbf{$\geq$5-hop} & \textbf{Total} & \textbf{1-hop}  & \textbf{2-hop} & \textbf{3-hop} & \textbf{$\geq$5-hop} & \textbf{Total}\\
      \midrule
      \multirow{5}{*}{EP-CP} & EEDP & \textbf{99.05} & \textbf{97.53} & \textbf{96.09} & \textbf{92.15} & \textbf{97.24} & \textbf{99.40} & \textbf{91.44} & \textbf{87.55} & \textbf{82.07} & \textbf{90.12}\\
      ~ & EEDP w/o $\mathcal{G}_{\text{adjlst}}$ & 95.90 & 91.38 & 91.56 & 87.53 & 92.66 & 95.50 & 86.09 & 87.10 & 85.21 & 88.48\\
      ~ & EEDP w/o $Path_{\text{EEDP}}$ & 98.40 & 95.12 & 82.22 & 55.53 & 89.62 & 98.61 & 90.09 & 69.41 & 47.52 & 76.41\\
      ~ & EEDP w/o \{$\mathcal{G}_{\text{adjlst}}$ + $Path_{\text{DAG}}$\} & 95.85 & 89.96 & 90.24 & 86.72 & 91.80 & 94.73 & 72.36 & 68.72 & 52.27 & 72.02\\
      \midrule
      \multirow{5}{*}{EP-DP} & EEDP & \textbf{92.30} & \textbf{78.54} & \textbf{79.50} & \textbf{69.01} & \textbf{82.74} & \textbf{95.42} & \textbf{61.72} & \textbf{65.12} & \textbf{47.72} & \textbf{67.50}\\
      ~ & EEDP w/o $\mathcal{G}_{\text{adjlst}}$ & 68.80 & 28.05 & 30.89 & 18.11 & 42.31 & 88.62 & 43.95 & 41.89 & 26.48 & 50.24\\
      ~ & EEDP w/o $Path_{\text{EEDP}}$ & 86.40 & 68.50 & 57.53 & 33.20 & 68.94 & 90.55 & 60.05 & 49.98 & 17.58 & 54.54\\
      ~ & EEDP w/o \{$\mathcal{G}_{\text{adjlst}}$ + $Path_{\text{DAG}}$\} & 70.50 & 30.35 & 31.73 & 21.53 & 43.89 & 85.97 & 43.02 & 37.92 & 19.43 & 46.59\\
      \bottomrule
    \end{tabular*}
    \end{threeparttable}
    \label{tab4}
\end{table*}

1. Overall, the proposed EEDP method outperforms all baseline methods. In short-distance scenarios, it surpasses the majority of baseline methods in both tasks across two datasets. In long-distance scenarios, EEDP achieves even greater improvements, demonstrating its effectiveness.

2. LLMs possess basic reasoning capabilities for graphs, but this ability diminishes as the reasoning path length increases. In short-distance scenarios, the reasoning performance of baseline methods is better than random. However, in long-distance scenarios, the performance of most baseline methods is worse than random. EEDP, however, outperforms random in all scenarios.

\subsection{Ablation Study}

To investigate the contribution of each component of the EEDP method to reasoning performance, we conducted ablation experiments with several simplified models: removing only the adjacency list $\mathcal{G}_{\text{adjlst}}$, removing all 'main backbone paths' $Path_{\text{EEDP}}$, and removing both the adjacency list $\mathcal{G}_{\text{adjlst}}$ and part of the 'main backbone paths' $Path_{\text{DAG}}$ shared by $\mathcal{G}$ and $DAG_{\text{EEDP}}$. Table \uppercase\expandafter{\romannumeral4} shows the performance variations of the EEDP method after removing different components.

Key observations are as follows:

1. Each component of EEDP is effective, particularly in long-distance scenarios. EEDP consistently achieves the highest accuracy. Removing any single component degrades performance, demonstrating the effectiveness of our human cognition-centric strategy.

2. EEDP exhibits robustness to changes in inference length. The performance reduction the EEDP method remains relatively smooth as inference distance varies. Methods including $Path_{\text{EEDP}}$ show a similar trend, indicating excellent stability.

\section{Conclusion}
We propose an end-to-end graph flattening method, End-to-End DAG-Path prompting (EEDP). Our EEDP method draws on the human cognitive process for graph data, optimizing the graph flattening process. We conduct experiments on our proposed dataset Merged\_1000 and ZINC\_test\_2500. The experimental results show that the EEDP method outperforms all baseline methods in zero-shot prediction scenarios for pure graph structures, demonstrating excellent inference performance in both short and long-distance contexts and robustness to varying connection distances.

\section*{Acknowledgment}
This research was partially supported by grants from the
National Key Research and Development Program of China
(2022YFC3303504), the University Synergy Innovation Program of Anhui Province (GXXT-2022-042), and the Fundamental Research Funds for the Central Universities (WK2150110034).



\end{document}